# Neural Machine Translation for Cebuano to Tagalog with Subword Unit Translation

Kristine Mae M. Adlaon[1,2], Nelson Marcos[1]
[1]College of Computer Studies, De La Salle University, Taft Avenue, Manila, Philippines
[2]Information Technology Education Program, University of the Immaculate Conception, Father Selga St., Davao City, Philippines
[1]{kristine_adlaon},{nelson.marcos}@dlsu.edu.ph, [2]{kadlaon}@uic.edu.ph

*Abstract*—The Philippines is an archipelago composed of 7, 641 different islands with more than 150 different languages. This linguistic differences and diversity, though may be seen as a beautiful feature, have contributed to the difficulty in the promotion of educational and cultural development of different domains in the country. An effective machine translation system solely dedicated to cater Philippine languages will surely help bridge this gap. In this research work, a never before applied approach for language translation to a Philippine language was used for a Cebuano to Tagalog translator. A Recurrent Neural Network was used to implement the translator using OpenNMT sequence modeling tool in TensorFlow. The performance of the translation was evaluated using the BLEU Score metric. For the Cebuano to Tagalog translation, BLEU produced a score of 20.01. A subword unit translation for verbs and copyable approach was performed where commonly seen mistranslated words from the source to the target were corrected. The BLEU score increased to 22.87. Though slightly higher, this score still indicates that the translation is somehow understandable but is not yet considered as a good translation.

## I. INTRODUCTION

The Philippines, being an archipelago composed of 7, 641 islands with around 150 different languages or dialects have been tagged to be linguistically diverse. These linguistic differences and regional divisions have created major difficulties in promoting educational and cultural development in the country. Although, this diversity has been unified by the declaration of a Philippine National Language which is Filipino (known before as Tagalog), the ability to effectively understand one another is of great importance.

Machine Translation (MT) is perhaps the most substantial way in which computers could aid human to human communication and even human to machine communication [6]. In the Philippines, much of the efforts (discussed further in the next section of this paper) in building a language translation system requires the development of different language resources and language tools which makes the task of translation difficult and tedious.

Neural machine translation has recently shown impressive results [7] [2] [11]. Originally developed using pure sequence-to-sequence models and improved upon using attention-based variants; NMT has now become a widely-applied technique for machine translation, as well as an effective approach for other related NLP tasks such as dialogue, parsing, and summarization.

In this paper, the researchers implemented and examined the result of using an NMT approach in language translation for Cebuano to Tagalog. More specifically, a recurrent neural network was used to perform the task of neural machine translation. These two languages (Cebuano and Tagalog) are the major languages widely spoken in the Philippines out of 150 other Philippine languages. To the researchers best knowledge, this work provides a ground-breaking result in using a never before applied approach in any machine translation research effort in the Philippines for Cebuano to Tagalog. The success of this research work will be a starting point for more machine translation work of other Philippine languages.

## II. RELATED WORKS

### A. Existing MT Systems and Approaches

Transfer-based English to Filipino MT system was designed and implemented using the functional reference grammar (LFG) as its formalism. It involves morphological and syntactical analyses, transfer, and generation stages. The whole translation process involves only one sentence at a time [4]. Another transfer-based system is ISAWIKA that uses ATN (Augmented Transition Network) as the grammar formalism. It translates simple English sentences into equivalent Filipino sentences at the syntactic level [9]. Fat [5] worked on a bilingual machine translation system designed for Tagalog and Cebuano. It exploits structural similarities of the Philippine languages Tagalog and Cebuano, and handles the free word order languages. It translates at the syntactic level only and uses Tagalog to Cebuano dictionary as the dataset. It does not employ morphological analysis in the system. This has been supplemented by the study of where a machine translation system for Tagalog and Cebuano was developed which focuses more on the morphological (analysis and synthesis) aspect of the two languages, taking into consideration their similarities and differences.

A more recent work used a statistical machine translation (SMT) approach for a bidirectional Filipino to English MT named as the ASEANMT-Phil. The system has experimented on different settings producing the

BLEU score of 32.71 for Filipino to English and 31.15 for English to Filipino [8]. Just like ASEANMT-Phil, another Moses-based SMT called the FEBSMT, incorporates periodic user feedback as a mechanism that allows the SMT to adapt to prevailing translation preferences for commonly queried phrases, and assimilate new vocabulary elements in recognition of the dynamically changing nature of the languages [1]. However, the evaluation scores were observed to decrease gradually as soon as it reached its peak due to the significant effect of the probability scores of the Language model and phrase tables affecting the translations of the baseline system that are correct, to begin with.

### B. . Neural Machine Translation Tools

Currently, there are several existing NMT implementations. Many systems such as those developed in the industry by *Google*, *Microsoft*, and *Baidu*, are closed source, and are unlikely to be released with unrestricted licenses. Many other systems such as *GroundHog*, *Blocks*, *tensorflowseq2seq*, *lamtram*, and *seq2seq-attn*, exist mostly as research code. These libraries provide important functionality but minimal support to production users. Perhaps a promising tool is the University of Edinburgh's *Nematus* system originally based on NYU's NMT system. Nematus provides high-accuracy translation, many options, clear documentation, and has been used in several successful research projects. This research explored the use of OpenNMT[1] system, an open-source framework for neural machine translation.

### III. RECURRENT NEURAL NETWORK for NMT

The idea behind RNNs is to make use of sequential information. In a traditional neural network, it is assumed that all inputs (and outputs) are independent of each other which are somehow not fitting for most of NLP tasks (e.g. machine translation) where proper ordering of words is very important. If one would want to predict the next word in a sentence or target words in the case of language translation, it is very important to know what words came before it.

RNNs are called recurrent because they perform the same task for every element of a sequence, with the output being dependent on the previous computations. Another way to think about RNNs is that they have a "memory" which captures information about what has been calculated so far. Fig 1 shows what a typical RNN looks like.

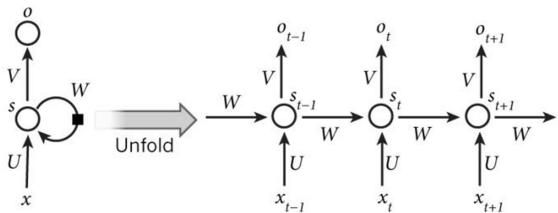

Figure 1. A recurrent neural network and the unfolding in time of the computation involved in its forward computation

Figure 1 shows a RNN being unrolled (or unfolded). By unrolling it means that the network is being written in a complete sequence. $X_1$ is the input at time step $t$. For example, $X_1$ is a one-hot vector corresponding to the second word of a sentence. $S_t$ on the other hand is the hidden state at time step $t$. It's the "memory" of the network. $S_t$ is calculated based on the previous hidden state and the input at the current step:

$$S_t = f\ (U_{X_t}\ +\ W_{S_{t-1}}\ ) \qquad (1)$$

Here, *ReLU* was used as the nonlinearity formula for the function *f*. While $s - 1$, which is required to calculate the first hidden state, was initialized to all zeroes. Lastly, $o_t$ is the output at step $t$. For example, predicting the next word in a sentence would be a vector of probabilities across the vocabulary.

Just like in language modeling, machine translation accepts a sequence of words as input in the source language (e.g. Cebuano) and output it as another sequence of words in the target language (e.g. Tagalog). The key difference is that, the output only starts after the complete input was seen because the first word of the translated sentences may require information captured from the complete input sequence.

### IV. METHODOLOGY

#### A. Data Collection

The data that was used in this research work came from a web resource[2] that contains a monolingual corpus of different languages created from translations of the Bible which is publicly available. Although it is stated in the source that the corpus Book, Chapter, and Verse indices are already aligned at a sentence level, the hand-crafted used in this work corpus was still double-checked manually for correct sentence alignment. The corpus is in XML format. Table I shows the general information of the dataset:

TABLE I. GENERAL INFORMATION OF THE DATASET

| Language | Tagalog | Cebuano |
|---|---|---|
| Family | Austronesian | Austronesian |
| Genus | Malayo-Polynesian | Malayo-Polynesian |
| Subgenus | Philippines | Philippines |
| No. of Speakers | 23,900,000 | 15,800,000 |
| Parts (New or Old Testament) | Complete | Complete |

#### B. Data Cleaning and Transformation

Data cleaning and transformation was performed in the dataset. This stage is important in order to transform the data into a format that can be used for processing.
1) Converted the dataset from an XML (.xml) file to a text (.txt) file.

---

[1] http://opennmt.net/

[2] http://christos-c.com/bible/

TABLE II. SAMPLE TRANSLATED SENTENCES WITH CORRESPONDING BLEU SCORE

| SP | Source | Target | Machine Translation | BLEU Score |
|---|---|---|---|---|
| 1 | ug miingon ang dios : magpaturok ang yuta ug balili , talamnon nga magahatag ug binhi; ug himunga nga mga kahoy nga magahatag ug bunga ingon sa ilang matang diin anaa kaniya ang iyang binhi sa ibabaw sa yuta : ug nahimo kini . | at sinabi ng dios , *sibulan* ang lupang *damo* , *pananim* na *nagkakabinhi* , at *punong* kahoy na *namumunga ayon* sa *kaniyang pagkakahoy* , na *taglay* ang *kaniyang* binhi sa *ibabaw* ng lupa , at *nagkagayon* . | at sinabi ng dios , *purihin* ang lupa , at *in iibig* ang lupa , na *tinimplahan* ng kapah *ingahan* , at *tatalab* sa *kaniya* ang *mga* kahoy na *iyon* , na *parang bunga* ng kanilang binhi : at *magiging banal* sa *kaniya* . | 16.52 |
| 2 | ug ang yuta nagpaturok ug balili , talamnon nga nagahatag ug binhi , ingon sa ilang matang; ug mga kahoy nga nagahatag ug bunga , nga maoy binhi niini , ingon sa ilang matang , ug nakita sa dios nga kini maayo . | at ang *lupa* ay *sinibulan* ng *damo* , *pananim* na *nagkakabinhi* , ayon sa *kaniyang pagkapananim* , at ng *punong* kahoy na *namumunga* , na *taglay* ang *kaniyang* binhi , ayon sa *kaniyang pagkakahoy* , at nakita ng dios na mabuti | at ang *lupain* ay *mapasuko* , at *kumapal* na *parang* binhi ng *pakikiapid* , na *gaya* ng kanilang *ama :* at ang *mga* kahoy na *cedro* , ay ayon sa *kanilang kapanganakan* , at nakita ng dios na mabuti . | 24.08 |
| 3 | ug dihay kahaponon ug dihay kabuntagon , adlaw nga ikatolo . | at *nagkahapon* at *nagkaumaga ang ikatlong* araw . | at *may malakas na hiyawan sa* araw *na yaon* . | 5.52 |

2) Removed all XML tags and Non-Printable ASCII Characters were removed.
3) Transformed all text to lowercase.
4) Removed repetitive sentences mostly in the Cebuano dataset.
5) Used only the Book of Genesis in Cebuano and Tagalog versions.
6) Aligned exact translations of sentences from the source to the target language manually.

After performing the different steps for the cleaning and transformation including that of the alignment, a total of 6,510 sentence pairs were generated.

### C. Splitting the Parallel Corpus

The parallel corpus was randomly split into training set, validation set, and test set. A total of 1, 220 (4%) sentence pairs were used for the test set (610) and validation set (610). The training set was oversampled (generated 10 times) 10 times after the split to ensure that the network has enough number of training sentence pairs to learn. Weights on the network are adjusted based on this set. A trained model is produced after this phase. After oversampling there were a total of 52,900 sentence pairs for the training set. The difference of the source input and the target output is measured using the validation set. Then, the test set is used to test whether the trained model is able to work on source and target pairs that are not seen during the training. Table II shows an example sentence pair where Cebuano is the source sentence and Tagalog is the target sentence.

TABLE III. SAMPLE CEBUANO TO TAGALOG SENTENCE PAIRS

| Cebuano | Tagalog |
|---|---|
| ug mitubag si jose kang faraon : ang damgo ni faraon usa lamang : ang dios nagpahayag kang faraon sa haduol na niyang pagabuhaton . | at sinabi ni jose kay faraon , ang panaginip ni faraon ay iisa; ang gagawin ng dios ay ipinahayag kay faraon : |
| ang pito ka mga vaca nga maanindot mao ang pito ka tuig ; ug ang mga uhay nga maanindot mao ang pito ka tuig : ang damgo usa lamang . | ang pitong bakang mabubuti ay pitong taon ; at ang pitong uhay na mabubuti ay pitong taon ; ang panaginip ay iisa . |

### D. Subword Unit Translation

Translation of rare words or unknown words is an open problem faced by NMT models as the vocabulary of these models are typically limited to some number [10]. A large proportion of unknown words are names, which can just be copied into the target text if both languages share an alphabet. If alphabets differ, transliteration is required (Durrani et al., 2014).

In this work, subword unit translation was performed on names and verbs; names (e.g. moises, jehova, dios) being *unknown words* and verbs *having complex morphological structure* (words containing multiple morphemes, for instance formed via compounding, affixation, or inflection).

### V. RESULTS AND DISCUSSION

In this section, results of training Cebuano to Tagalog sentence pairs in a Recurrent Neural Network (RNN) with subword translation are shown. Two experiments were performed: (1) the RNN model was fed with sentence pairs without subword translation, and (2) the RNN model was fed with sentence pairs with subword translation.

### A. Without Subword Translation

After splitting the corpus, training was then performed. Logarithmic loss values generated during the training to measure the performance of the model on how far the source values are from the target indicated values were extracted. The ideal loss value is 0. Training was stop at step 25000 (loss = 1.39) when loss was consistently below 2 starting step 21000 (loss = 2). Figure 2 shows the loss values over 25000 training iteration.

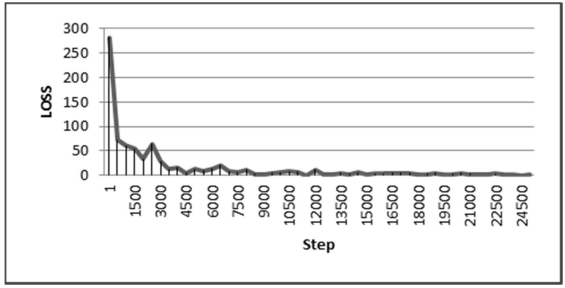

Figure 2. Loss values over training step time of the model.

After the training, translation was evaluated using the test cases via the BLEU metric. Result of the evaluation is a BLEU score indicating that the higher the BLEU score the better the translation. Fig 3 clearly shows that the longer we train the model the translation is getting better and better. To avoid overfitting, a threshold of loss = 2 was chosen as a stopping parameter.

The model achieved a BLEU score of 20.01 for Cebuano to Tagalog translation without the word and subword translation. This score indicates a somewhat understandable translation. Table III shows a sample translated sentences taken from the test set. The words that are strongly emphasized and italicized are the words not found in the translation.

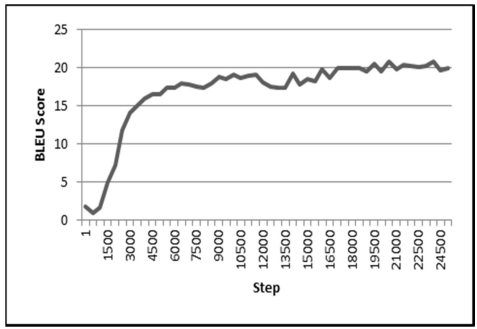

Figure 3. BLEU Scores over step time of the model.

Looking at the sentence pairs shown in Table III, the translation seems to have a much better BLEU score result on longer sentences compared to short sentences. This is because unigram correct translation is much higher on longer sentences. Although this is not true in all sentence pairs seen on the test set. An example would be the translation for this source sentence:

*ug ang babaye mitubag sa bitin : makakaon kami sa bunga sa mga kahoy sa tanaman :*

The correct translation should be:

*at sinabi ng babae sa ahas , sa bunga ng mga punong kahoy sa halamanan ay makakakain kami:*

The machine translated it as:

*at sinabi ng babae sa ahas , kami ay tumakas sa bunga ng mga punong kahoy ,*

This sentence pair got a high BLEU score of 58.07. In the translation table result shown in Table III, it can be observed that in sentence pair 2 there is an insertion of the word *cedro* (a noun based on context) which is never seen in the source sentence. When the word was searched for the word cedro in the whole training set, we found that the word *cedro* is a name of a tree which was always mentioned in the training set as *kahoy na cedro*. Looking at the sentence pair 2 in the test set, where punong kahoy was mentioned, the machine translated it as *kahoy na cedro*.

Another observation is that we have seen mistranslations for names of people in the Bible which we think are very critical to address since the Bible has a lot of important biblical names. The following are samples that were seen as incorrect translations:

TABLE IV. CASE OF OUT-OF-VOCABULARY WORDS

| Source | Target | Translation |
|---|---|---|
| abimelech | abimelech | moises |
| eschol | eschol | elisur |
| mamre | mamre | jacob |
| sarai | sarai | moises |
| abram | abram | elisabeth |

Further investigation to this case has led us to the conclusion that this is a case of out-of-vocabulary (oov) word problem. The words in the source are words that are not found in the training set. Even a single instance of the word did not exist in the training set.

In this case, oovs were handled by replacing it with a word (still a name of a person) commonly found in the training set. This is the very nature of sequence modelling using recurrent neural network. The network was able to determine that these unknown words are names of person because of its preceding words. Other common words that were handled properly during translation are the the words *dios, abraham, panginoon, ismael*. A simple copyable approach of these words to the translation may not be the best solution since there are instances as well that the source word is different from the target word such as the word *jehovah* in Cebuano is *panginoon* in Tagalog. To handle this case, a word unit translation copying the source word to the target word was applied (discussed further in the next experiment).

### B. With Subword Unit Translation

Before the dataset was fed to the model for translation, a word and subword translation was performed to the parallel corpus from Cebuano to Tagalog. In the experiment, the words that have the highest frequency values were extracted from the vocabulary. With reference to the statement of [3], that a large proportion of unknown words are names, the researchers identified highest frequency of names and manually examined its equivalent translation in the target corpus. An example of this is shown in Table V.

TABLE VI. COMPARISON OF TRANSLATED SENTENCES WITH AND WITHOUT SUBWORD TRANSLATION

| SP | Reference | Result without Subword Translation | BLEU Score | Result with Subword Translation | BLEU Score |
|---|---|---|---|---|---|
| 1 | at sinabi ng dios , *sibulan* ang lupang *damo* , *pananim* na *nagkakabinhi* , at *punong* kahoy na *namumunga ayon* sa *kaniyang pagkakahoy* , na *taglay* ang *kaniyang* binhi sa *ibabaw* ng lupa , at *nagkagayon* . | at sinabi ng dios , *purihin* ang lupa , at *iniibig* ang lupa , na *tinimplahan* ng *kapahingahan* , at *tatalab* sa *kaniya* ang *mga* kahoy na *iyon* , na *parang bunga* ng *kanilang* binhi : at *magiging banal* sa *kaniya* . | 16.52 | at sinabi ng dios , *mapayapa nawa* ang *lupain* , at *iginuhit* sa *iyo* , ng *kasagutan* at ng *laryo* , na *mga* kahoy na *itinutubo* ng kahoy sa lupain ; at *siya'y naging parang pakinabang* sa ibabaw ng lupa . | 18.97 |
| 2 | at ang *lupa* ay *sinibulan* ng *damo* , *pananim* na *nagkakabinhi* , ayon sa *kaniyang pagkapananim* , at ng *punong*kahoy na *namumunga* , na *taglay* ang *kaniyang* binhi , ayon sa *kaniyang pagkakahoy* , at nakita ng dios na mabuti | at ang *lupain* ay *mapasuko* , at *kumapal* na *parang* binhi ng *pakikiapid* , na *gaya* ng *kanilang ama :* at ang *mga* kahoy na *cedro* , ay ayon sa *kanilang kapanganakan* , at nakita ng dios na mabuti . | 24.08 | at ang *lupain* ay *mapasuko* at *nilakad* , , na *mainam* na *harina* , at ang binhi ng *pakikiapid* , at ang *mga* kahoy na *itinutubo* ng kahoy na *cedro* , ay *magtataglay* ng *kanilang ulo* at ng dios na mabuti . | 15.05 |
| 3 | at *nagkahapon* at *nagkaumaga ang ikatlong* araw . | at *may malakas na hiyawan sa* araw *na yaon* . | 5.52 | at nagkahapon at nagkaumaga ang ikatlong araw . | 100.00 |

TABLE V. SAMPLE NAMES EXTRACTED FROM THE DATASET HAVING HIGH FREQUENCY

| Names | Possible Translations | Replaced with | # of Occurences |
|---|---|---|---|
| jehova | jehova, panginoon, dios none | jehova | 1580 |
| dios | jehova, panginoon, dios none | dios | 622 |
| israel | katilingban, israelita, israelihanon, none | israel | 665 |
| moises | moises, siya, kaniya, none | moises | 651 |
| aaron | aarong | aaron | 292 |

It can be observed in Table V, that both the words *jehova* and *dios* can be used to translate one from the other. There are instances in the dataset that the word *jehova* from the source is referred to as *dios* or *panginoon*. The none tag in the possible translation column is a case where in there is no equivalent translation of the word in the target language given a sentence pair. It can be noticed as well that most of the possible translations for the name moises are pronouns.

Word translation was performed for most frequently occurring names that were mistranslated. The researchers applied the copyable approach for word translation in names. Example, for the word *dios* it was found out that the most frequent translation of the word *dios* is also *dios*, therefore any instance that the word *dios* is found in the source language and is translated as *panginoon* or *jehova*; it will be replaced with the word *dios*. If none, the word *dios* will be inserted in the target language.

Aside from names, most frequently occurring verbs found in the dataset were examined. Table VII shows 5 of the most frequently occurring verbs. Subword unit translation was performed in these words. Example, the word *ngadto* has 11 possible translations. After inspection it was found out that the most frequently used translation of the word *ngadto* is *paroon*. After knowing that the translation is p*aroon*, all occurrences of the word *ngadto* in the source language was replaced with the word *paroon* in the target language.

TABLE VII. SAMPLE LIST OF WORDS AND ITS POSSIBLE TRANSLATIONS

| Verbs | Possible Translations | # of Occurences |
|---|---|---|
| ngadto | pumaroon, bumaba, pasubasob, none, magsisiyaon, nagsibaba, napasa, mula, paroon, yumaon, papasasa | 541 |
| halad | handog, hain, kaloob | 169 |
| sumala | ayon, gaya ng, kung paanong | 273 |
| ingon | sinabi, gaya nito | 883 |
| uban | kasalamuha, none | 332 |

After doing subword translation, logarithmic training loss values generated during the training to measure the performance of the model on how far the source values are from the target indicated values were extracted. The ideal loss value is 0. Training was stop at step 9000 (loss = 1.22) when loss was consistently below 2. Figure 2 shows the loss values over 9000 training iterations.

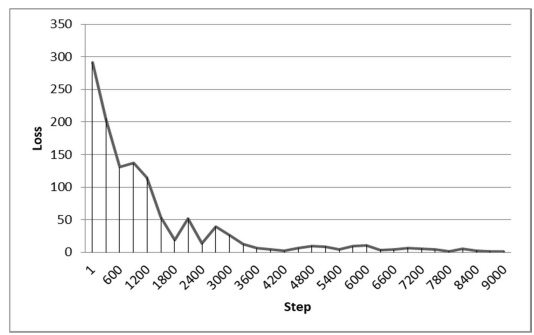

Figure 4. Loss values over training step time of the model.

After the training, translation was evaluated using the test cases via the BLEU metric. The model achieved a slightly higher BLEU score of **22.87** for Cebuano to

Tagalog translation with subword translation. Table VI shows that the impact of the correction (Subword translation) is greatly seen on sentence pair number 3. A perfect translation of the reference sentence was outputted. On the other hand it did not performed well on sentence pair number 2. The BLEU score got even lower.

## VI. CONCLUSION

In this paper, a never before applied approach for language translation was applied for a Cebuano to Tagalog translator. A Recurrent Neural Network was used to implement the translator using OpenNMT framework. The performance of the translation was evaluated using the BLEU Score metric. A subword unit translation was performed. Two experiments were conducted to check the effect of subword correction to the translation performance. For the Cebuano to Tagalog translation, BLEU produced a score of 20.01 without the subword correction. This score indicates that the translation is somehow understandable but is not yet considered as good translation. This low score maybe because some words on the test data are out of the vocabulary or these words are not seen in the training data. Common and much more obvious mistranslations are those that of the person's name. A simple copyable approach of these words to the translation was applied, although may not be the best solution since there are instances as well that the source word is different from the target word such as the word *Jehovah* in Cebuano and *Panginoon* in Tagalog.

Another observation is that, since BLEU score metric of evaluation evaluates the translation on an exact word match bases no point is given to translations that are subword of the target word. For example, the word *pangalang* and *pangalan* in the sentence pair;

Target:  kaya ang *pangalang* itinawag ay babel;
MT:      kaya't ang *pangalan* ng panginoon ay tinawag a kibroth-hattaavah;

We performed subword translation for verbs since verbs are the ones that are seen to have morphological complexities. For this experiment, verbs that occurred most frequently in the corpus were extracted. After extraction, manual correction of the mistranslated words was performed. After training, translation was again evaluated via getting the BLEU score. Although, there was a slight increase in the translation performce with a score of 22.87, this still indicates a somehow understandable translation.

For future work, it is recommended to increase the number of training pairs to improve the translation performance. Also, consider including another domain of dataset aside from the bible. A more in-depth study on how to handle out-of-vocabulary words should also be taken into consideration.